\documentclass[sigconf]{acmart}
\AtBeginDocument{
  }

\setcopyright{acmlicensed}
\copyrightyear{2026}
\acmYear{2026}
\acmDOI{XXXXXXX.XXXXXXX}
\acmConference[KDD '26]{the 32nd ACM SIGKDD Conference on Knowledge Discovery and Data Mining}{August 9--13,
  2026}{Jeju, Korea}

\acmISBN{978-1-4503-XXXX-X/2026/08}

\begin{document}

\title{Student Flow Modeling for School Decongestion via Stochastic Gravity Estimation and Constrained Spatial Allocation}

\author{Sebastian Felipe R. Bundoc}
\authornote{Both authors contributed equally to this research.}
\email{ecair.sbundoc@deped.gov.ph}
\orcid{0009-0007-1973-840X}
\author{Paula Joy B. Martinez}
\authornotemark[1]
\email{ecair.pmartinez@deped.gov.ph}
\affiliation{
  \institution{Center for AI Research}
  \institution{Department of Education}
  \country{Philippines, PH}
}

\author{Sebastian C. Ibañez}
\affiliation{
  \institution{Center for AI Research}
  \institution{Department of Education}
  \country{Philippines, PH}}
\email{ecair.sibanez@deped.gov.ph}

\author{Erika Fille T. Legara}
\affiliation{
  \institution{Center for AI Research}
  \institution{Department of Education}
  \country{Philippines, PH}}
\email{ecair.elegara@deped.gov.ph}

\renewcommand{\shortauthors}{Bundoc \& Martinez et al.}

\begin{abstract}
School congestion, where student enrollment exceeds school capacity, remains a major challenge in low- and middle-income countries.
It highly impacts learning outcomes and deepens inequities in education.
While education subsidy programs that transfer students from public to private schools offer a mechanism to alleviate congestion without capital-intensive construction, they often underperform due to fragmented data systems that hinder effective implementation.
The Philippine Educational Service Contracting program, one of the world's largest educational subsidy programs, exemplifies these challenges, falling short of its goal to decongest public schools.
This data fragmentation prevents the science-based and data-driven analyses needed to understand what shapes student enrollment flows, particularly how families respond to economic incentives and spatial constraints.
In this work, we introduce a computational framework for modeling student flow patterns and simulating policy scenarios.
By synthesizing heterogeneous government data across nearly 3,000 institutions, including school geolocations, infrastructure records, enrollment statistics, school tuition fees, student flows, road networks, subsidy amounts and slots, we employ a stochastic gravity model estimated via negative binomial regression to derive behavioral elasticities for distance, net tuition cost, and socioeconomic determinants.
These elasticities inform a doubly constrained spatial allocation mechanism that simulates student redistribution under varying subsidy amounts while respecting both origin candidate pools and destination slot capacities.
We find that geographic proximity constrains school choice four times more strongly than tuition cost, and that slot capacity–not subsidy amounts—is the binding constraint.
Our work demonstrates that subsidy programs alone cannot resolve systemic overcrowding, and computational modeling can empower education policymakers to make equitable, data-driven decisions by revealing the structural constraints that shape effective resource allocation, even when resources are limited.
\end{abstract}




\maketitle

\section{Introduction}

School congestion, defined as enrollment exceeding a school's physical and pedagogical capacity, remains a persistent barrier to global education quality \cite{barrett:2018, carro:2023}.
In Lower- and Middle-Income Countries (LMICs), this challenge is particularly acute, with student-teacher ratios often doubling global averages \cite{unesco:2022}.
In the Philippines, over 5.1 million students are enrolled in overcrowded classrooms \cite{edcom:year2}.
To address this, the Department of Education utilizes the Educational Service Contracting (ESC) program—one of the world's largest public-private partnerships in education~\cite{worldbank:2011}—to provide tuition subsidies that incentivize the transition of Grade 7 students from overextended public schools to participating private institutions~\cite{egastpe:2017} nationwide.

Despite the program's scale, the Department's internal capacity to measure its impact on systemic decongestion has historically been limited.
To resolve this, our team, together with the Government Assistance and Subsidies Services Office, was tasked with transforming how the Department understands and relieves public school overcrowding.
This paper documents the computational policy framework that we developed to address the systemic barriers the Department of Education encounters.
Through our experience in administering the ESC, it became evident that our primary obstacles were fragmented information systems and analytical frameworks that failed to account for necessary complexity.
First, our critical administrative data on school capacity, spatial accessibility, and student mobility were siloed across disparate bureaus and other national government agencies, preventing a unified view of the system.
Second, traditional policy analysis relied on linear, aggregate models that failed to capture the nonlinear interactions between student flows, spatial constraints, and economic incentives.

Our computational framework integrates heterogeneous administrative data from nearly 8,000 schools with open-source geospatial data and government data from other national government agencies to construct a comprehensive representation of student flows in the Philippine basic education system.
Our contribution lies in using a stochastic gravity-based model to implement counterfactual decongestion and perform policy simulations.
By framing student mobility within a gravity approach, we characterize the interaction between origins and destinations as a function of mass (school capacity and candidate beneficiaries) and friction (e.g., road distance and net tuition costs).
These parameters are then integrated into a doubly constrained spatial allocation mechanism that simulates how student flows shift across the network when we adjust subsidy amounts.

Our work makes three primary contributions to the science of educational resource allocation.
First, we demonstrate how harmonizing fragmented administrative datasets with open-source data and records from other national government agencies enables system-wide visibility into student flow networks that were previously opaque to both policymakers and researchers.
This yields the first origin-destination flow graph of the Philippine basic education system at national scale.
Second, through a gravity-based spatial interaction model, we establish an empirical characterization of school choice behavior among subsidy-eligible families.
Geographic accessibility dominates economic cost by a factor of four in determining enrollment flows, a finding that challenges the prevailing policy assumption that subsidy increases are the primary lever for driving private school participation.
Third, by coupling this behavioral estimation with a doubly constrained stochastic allocation framework, we uncover structural properties of the system that aggregate statistics alone cannot reveal.
We show that ESC enrollment is capacity-bound rather than price-bound, and that over half of the system's untapped decongestion potential lies in latent origin-destination pathways not yet observed in historical data.
We apply our computational framework to the three most populous regions of the Philippines, the National Capital Region, Region III, and Region IV-A, which together account for the vast majority of the country's school congestion.

Beyond these empirical discoveries, this work demonstrates how computational methods from spatial science and AI can make the structure of a national educational system accessible to scientific inquiry.
Where the Department of Education previously relied on descriptive reporting of historical enrollment, our framework enables prospective simulation of counterfactual policy scenarios.
This reveals not only how students would redistribute under alternative subsidy regimes but, more fundamentally, identifies the binding constraints on the system's decongestion potential: physical school capacity rather than tuition cost.
While developed for the Philippine basic education context, the methodology generalizes to spatial allocation problems in which behavioral models of individual choice must be reconciled with physical capacity constraints across a network of facilities, including healthcare access, public transit planning, and social service delivery.

\section{Related Literature}

\subsection{Philippine Educational Service Contracting}

The ESC program was established in 1998 to reduce public school congestion by providing tuition subsidies for elementary graduates to enroll in partner private secondary schools \cite{egastpe:2017}.
It has grown exponentially over the years, serving nearly one million students through collaboration with over 3,000 private schools by the 2017-2018 school year \cite{ocampo_lucasan:2019}.
However, despite the program's substantial scale, fundamental weaknesses in its design and governance are well-documented, particularly with respect to its two levers of control: subsidy slots and subsidy amount.
A World Bank \cite{worldbank:2011} evaluation discovered that the current distribution of grantees by region is not proportional with the estimated number of excess students in each region, resulting in some regions receiving disproportionately larger shares while others were underserved.
A local study also found issues with the distribution mechanism at the school level \cite{ocampo_lucasan:2019}.
With respect to subsidy amounts, \cite{brodeth:2025} revealed significant inadequacies in current grant amounts, which have remained unchanged since 2017 and are merely based on the geographic location of the participating private school—those located in the capital region receive the highest amount, while schools established in more provincial regions are granted the lowest.
While existing literature identifies the misallocation of slots and the inadequacy of subsidies, it often treats these factors as isolated administrative issues.
We argue that these two levers, subsidy slots and amounts, must be modeled as interdependent components of a redistribution mechanism that analyzes student flow patterns, identifies underutilized private sector capacity, and systematically reallocates students from congested public schools to available private seats.

\subsection{Modeling Student Flow}

We model student flows using a framework analogous to Newton's law of gravitation.
The gravity model posits that flows between origin $i$ and destination $j$ are proportional to origin "mass" and destination "mass" and inversely proportional to the distance between their centers of mass.
This framework has been widely used across diverse domains such as in transportation and urban planning~\cite{tsekeris:2006,lewer:2008,persyn:2015,simini:2021,zhang:2025}, as well as in education, particularly in understanding student migration patterns~\cite{bratti:2019,bacci:2020,kosztyn:2021,litmeyer:2023,pelegrini:2022}.
However, unlike physical gravity, which holds exactly, economic gravity models are approximations requiring stochastic specifications~\cite{silva:2006}.
Individual decisions depend on many unmeasured factors, including personal preferences, family connections, information availability, and random events that create deviations from theoretical predictions.
The stochastic gravity equation takes the form $F_{ij} = \alpha_0 O_i^{\alpha_1} D_j^{\alpha_2} d_{ij}^{\alpha_3} \eta_{ij}$, where $\eta_{ij}$ is a multiplicative error term with $E(\eta_{ij}|\text{regressors}) = 1$, assumed to be statistically independent of the regressors.

Traditionally, gravity models were estimated by ordinary least squares on log-transformed flows.
However,~\citet{silva:2006} demonstrate that heteroskedasticity renders OLS estimates biased and inconsistent, even after controlling for fixed effects.
They proposed the Poisson Pseudo-Maximum Likelihood estimation, which handles heteroskedasticity consistently and accommodates zero flows.
This approach has since become standard practice in spatial interaction modeling~\cite{larch:2025}.

Because student enrollment flows emerge from individual school choice decisions, the gravity model can also be viewed through the lens of microeconomic choice behavior.~\citet{mcfadden:1974} developed the seminal multinomial logit model based on random utility maximization, originally applied to travel behavior analysis but since widely adopted in education contexts~\cite{hastings:2008,abdulkadirolu:2020,awadelkarim:2023,guan:2025}.
Under this framework, the aggregate student counts we observe are the realized outcomes of individual utility-maximizing choices.
Following the mathematical framework established by~\cite{guimaraes:2003}, when the number of alternatives in a choice set is large, as is the case with thousands of origin-destination pairs in our dataset, a Poisson-based regression becomes mathematically equivalent to a conditional logit model.
This equivalence allows us to treat the gravity equation as a computationally tractable method to analyze individual choice behavior.

\subsection{Constrained Spatial Allocation}

The mathematical foundation for redistributing students across a network originates in the transportation problem of linear programming~\cite{hitchcock:1941,koopmans:1949}, which optimizes resource flows from supply points (origin schools) to demand points (destination schools) while minimizing a cost function, typically transportation costs.
Modern extensions of this logic combine machine learning-based discrete choice models with stochastic optimization to predict student enrollment under different zoning plans~\cite{guan:2025}, while others emphasize that reserving seats for specific assignment rules can reduce strategic misrepresentation and improve equity~\cite{xu:2023}.
Our algorithm follows this logic by treating the candidate pool at origin schools as the volume to be allocated and the unutilized slots at destination schools as the system's capacity constraints.
We extend the classical framework by using the OD-specific behavioral elasticities derived from the gravity model to weight these allocations, rather than assuming fixed or uniform preferences across all school pairs.

This redistribution mechanism aligns further with~\citet{wilson:1971}'s framework for doubly constrained spatial interaction.
Unlike \\production-constrained models that only fix total movers from an origin, doubly constrained models introduce balancing factors to ensure that destination totals do not exceed known capacities, represented in this paper as the available subsidy slots.
We then incorporate lexicographical searches to identify new, viable origin-destination pairs.
This approach is grounded in the literature on non-compensatory decision-making and student-optimal matching theories, which suggests that households do not weigh all school attributes simultaneously but rather filter options through a hierarchy of priorities, where distance is often the highest~\cite{awadelkarim:2023}.

\section{Methodology}

\begin{figure*}[t]
    \centering
    \includegraphics[width=\textwidth]{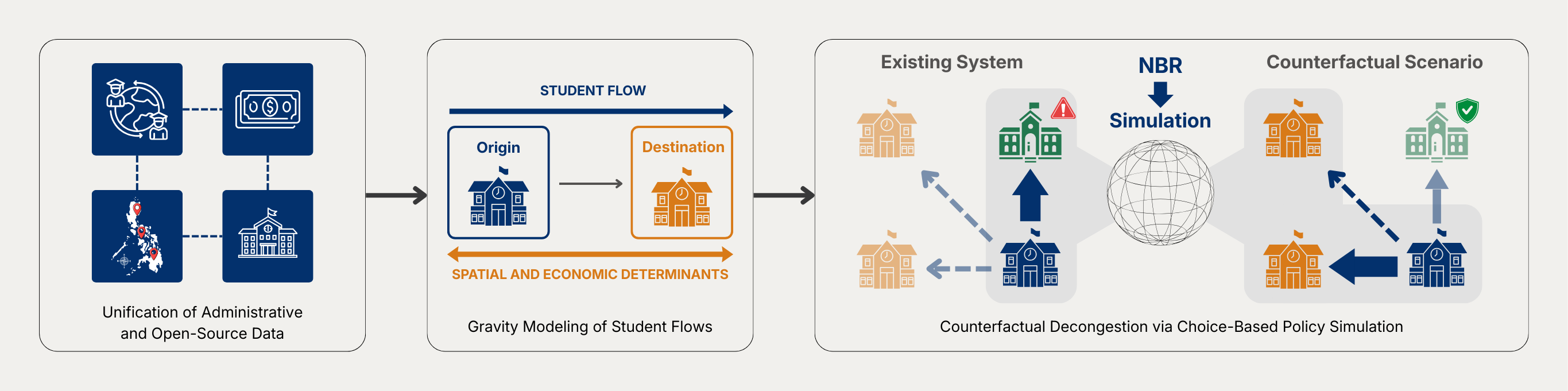}
    \caption{Overview of the computational policy framework for student redistribution. Institutional data from multiple national government agencies and open-source geospatial datasets are unified to represent the national school network. A stochastic gravity model estimated via negative binomial regression derives behavioral elasticities for distance, net tuition cost, and other determinants. Predicted flows are then allocated through a doubly constrained mechanism that iteratively assigns students until candidate pools or slot capacities are exhausted, enabling counterfactual evaluation of subsidy and slot allocation policies.}
    \label{fig:pipeline}
\end{figure*}

\subsection{Data Sources}

This study was conducted through an internal collaboration between the Center for AI Research and the Government Assistance and Subsidies Service Office of the Department of Education in the Philippines.
The datasets used span school years 2023-2024 and 2024-2025, and were sourced from multiple information systems within the Department.

\subsubsection{Public and private school coordinates}

School coordinates enable spatial analysis to identify nearby private school alternatives when redirecting incoming Grade 7 students away from overcrowded public high schools.
Public school coordinates are obtained from the national school registry, while private school coordinates are compiled from the administrative records of the country's 18 regional field offices.
Preprocessing standardizes coordinate formats across sources and filters out records lacking valid coordinates or school identifiers, yielding 47,821 public and 11,831 private school locations for geospatial matching.

\subsubsection{Inter-school road network distances}

Road network distances between schools are computed using the Open Source Routing Machine, a routing engine built on OpenStreetMap data.
OSRM's Multi-Level Dijkstra algorithm enables efficient computation of pairwise distances for all 13,106 schools within the three largest regions in the Philippines—the National Capital Region and two adjacent regions—yielding a distance matrix of over 171 million school pairs.
Unlike straight-line distances, road network distances account for actual travel paths, providing more realistic estimates of geographic accessibility for student redirection.

\subsubsection{School-level enrollment}

School-level enrollment data provides aggregate student counts for each school, disaggregated by grade level.
Sourced from the national enrollment database and spanning three school years, the dataset covers approximately 60,000 public and private schools annually.
High school enrollment figures are particularly critical: when compared against seating capacity, they reveal which public schools are operating beyond capacity and which private schools have room to absorb additional students.

\subsubsection{Public and private school seats data}

Seating capacity data quantifies the maximum number of students each school can physically accommodate.
Public school capacity is derived from official seat-learner ratio records, while private school capacity is compiled from classroom furniture inventories.
Schools where high school enrollment exceeds seating capacity are flagged as congested, with the excess enrollment termed "aisle learners" or students who lack dedicated seats.
This metric directly informs the choice and redirection models by quantifying overcrowding at each public school and the absorptive capacity at nearby private schools.

\subsubsection{ESC beneficiary records}

Beneficiary records from the Education Service Contracting (ESC) program identify students currently receiving tuition subsidies at participating private schools.
Spanning three school years and totaling approximately 2.7 million individual records, this dataset enables the model to distinguish between existing beneficiaries and non-beneficiaries within observed student flows.
Since beneficiaries are already enrolled in private schools through the subsidy program, only non-beneficiaries, or students flowing toward congested public schools, are candidates for redirection to private alternatives with unutilized subsidy slots.

\subsubsection{ESC program data}

Program-level data describes the resources and constraints governing the ESC subsidy.
Tuition fee records indicate what participating private schools charge, while mandated subsidy amounts specify the government contribution per beneficiary.
The gap between the two represents the out-of-pocket burden on families.
Slot allocation data specifies how many subsidized seats each school is authorized to fill.
Crucially, unutilized slot counts from official program records indicate remaining capacity at each school, serving as the binding constraint in the redistribution algorithm.
Finally, certification ratings from the national quality assurance system provide a measure of school performance, enabling the model to prioritize higher-quality alternatives when multiple redirection options exist.
These institutional quality assurance ratings were translated into hierarchical ordinal values to enable quantitative comparison of school quality during choice modeling and redirection prioritization.

\subsubsection{Municipality Income Data}

To account for the socioeconomic determinants of student flow, we integrated municipal-level income data from the Cities and Municipalities Competitiveness Index.
Published annually by the Department of Trade and Industry, this dataset provides a standardized framework for evaluating the financial capacity and economic dynamism of Local Government Units (LGUs).
We specifically utilized LGU income as a proxy for the socioeconomic status of both origin and destination municipalities.
Origin-side income data allows the model to identify the financial capacity of a student’s home municipality, while destination-side income serves as a proxy for the localized cost of living.

The student flow dataset is the primary analytical output of this study, synthesized by merging the administrative records described above.
By joining enrollment databases, beneficiary records, and school registries, we constructed a dataset of student transitions from Grade 6 to Grade 7.
These origin-destination (OD) records, spanning three school years and totaling approximately 2.7 million individual transitions, form the empirical foundation for our gravity modeling and redistribution simulation.

For each transition, we integrated the road network distance between the origin and destination, the net tuition cost (calculated as the difference between private school fees and the ESC subsidy), and the socioeconomic status of the origin municipality.
This unified view enables the model to differentiate between spatial interactions—geographic movements driven by proximity and historical feeder patterns—and subsidy-responsive flows incentivized by the ESC program.

\subsection{The Empirical Model of Student Flows}

We employ a stochastic gravity model to analyze student flows across the three largest regions in the Philippines.
Individuals are assumed to be rational agents who maximize their utility by evaluating the benefits of a destination against costs, such as travel costs implied through distance.
We describe student flows $F_{ij}$ by a general gravity model where the movement from origin $i$ to destination $j$ is a function of origin-side socioeconomic characteristics, destination-side attractiveness, and the spatial separation between them.
Traditionally, gravity models are estimated by ordinary least squares (OLS) on log-transformed flows.
However, the implications of Jensen's inequality render OLS estimates biased and inconsistent in the presence of heteroskedasticity, even after controlling for fixed effects \cite{silva:2006, larch:2025}.
As $E(\ln y) \neq \ln E(y)$, log-linearization of the gravity equation leads to systematic errors in estimating true elasticities.
While Poisson models are known to solve this, they only provide correct estimations if the equidispersion theorem applies.
Because student transitions exhibit a power law distribution, where a small number of OD pairs account for the majority of flows, the resulting overdispersion violates the equidispersion requirement of a Poisson model, necessitating the use of a Negative Binomial Regression (NBR).
A detailed comparison justifying the transition from OLS and Poisson to NBR is provided in Appendix~\ref{app:model_comparison}.

We formalize the stochastic gravity model as:

\begin{equation}
F_{i,j} = \alpha_0 O_i^{\alpha_1} D_j^{\alpha_2} d_{i,j}^{\alpha_3} c_{i,j}^{\alpha_4} \eta_{i,j}
\end{equation}

\begin{equation}
E(F_{i,j}|O_i, D_j, d_{i,j}, c_{i,j}) = \alpha_0 O_i^{\alpha_1} D_j^{\alpha_2} d_{i,j}^{\alpha_3} c_{i,j}^{\alpha_4}
\end{equation}

where $F_{i,j}$ is the flow of ESC beneficiaries from origin $i$ to school $j$, $O_i$ captures origin school characteristics, $D_j$ represents destination school attractiveness, $d_{i,j}$ is road network distance (km), $c_{i,j}$ is net enrollment cost (tuition minus ESC subsidy, in thousands of pesos), and $\eta_{i,j}$ is a multiplicative error term with $E(\eta_{i,j}|\text{regressors}) = 1$.

Following \cite{anderson:2003}, we also use fixed effects to control for all time-invariant origin and destination factors.
Our final equation utilizes a log-link function to maintain the multiplicative nature of the gravity framework while avoiding the biases inherent in manual log-linearization:

\begin{equation}
    \begin{split}
    \ln E(F_{i,j}) &= \beta_0 + \gamma_i + \delta_j \\
    &\quad + \alpha_3 \ln d_{i,j} + \alpha_4 \ln c_{i,j} \\
    &\quad + \beta_1 \ln \text{OriginIncome}_i + \beta_2 \ln \text{DestIncome}_j \\
    &\quad + \beta_3 \text{Rating}_j
    \end{split}
\end{equation}

Where $\beta_0$ is the intercept, $\gamma_i$ and $\delta_j$ represent regional fixed effects, and $\ln d_{i,j}$ and $\ln c_{i,j}$ are the log-transformed distance and net cost variables used to estimate the elasticities $\alpha_3$ and $\alpha_4$.
The iterative process for selecting these parameters is detailed in Appendix~\ref{app:model_specification}.
Furthermore, the dispersion parameter ($\alpha = 0.3948$) confirms the NBR as the robust choice for this dataset over the restrictive Poisson alternative.
To account for the potential intra-cluster correlation of student flows originating from the same school, we employ clustered robust standard errors at the origin school level.
This ensures that our inference remains valid in the presence of unobserved local factors that may drive correlated enrollment behaviors among students within the same geographic catchment area.

\subsection{Candidate Beneficiary Pool and Policy Simulation}

To assess the efficacy of varying policy interventions, we apply the estimated model parameters to simulate the behavioral response of students to changes in the subsidy amount.
The subsidy is treated as a direct reduction in the economic constraint $c_{ij}$, allowing us to model the sensitivity of non-beneficiary flows and predict the resulting student redistribution under different levels of financial support.

A critical component of this simulation is the definition of the $\text{candidate beneficiary pool}$ ($E_i$) for each origin school $i$.
This is the untapped cohort of potential subsidy beneficiaries.
While the model is trained on historical beneficiary flows, the policy objective is to project the movement of students who have not yet opted into the ESC program.
We define this candidate pool as:

\begin{equation}
    E_i = \text{Enrollment}_i - \sum_j \text{Beneficiaries}_{ij}
\end{equation}

$\text{Enrollment}_i$ is the total Grade 6 student population at origin school $i$, and the summation represents the existing flow of students into the private sector through the subsidy program.
This pool ensures that the simulation respects the physical population limits and avoids overestimating decongestion by constraining predicted flows such that $\sum_j \hat{y}_{ij} \leq E_i$.
To capture demand beyond historical observations, the simulation expands the set of possible school choices (OD pairs) for each student.
This augmented set includes not only established pathways but also counterfactual pairings identified through a lexicographical search of the school network.

\subsection{Counterfactual Decongestion via Choice-Based Policy Simulation}

The augmented candidate beneficiary pool from Section~3.3 provides, for each origin-destination pair $(i,j)$, a scenario-specific predicted flow $\hat{y}_{ij}^{(s)}$ representing the expected number of students from origin $i$ who would enroll at ESC school $j$ under policy scenario $s$.
These predictions are unconstrained: they reflect the behavioral response to changes in net cost $c_{ij}$ but do not account for finite school capacity or the competition between ESC destinations for the same origin pool.
We now introduce a simulation framework that imposes both supply- and demand-side constraints to produce realistic enrollment projections, and then attributes the resulting flows to public school decongestion.

\paragraph{Constrained Allocation under Coupled Depletion.}

Two capacity constraints bind simultaneously in the ESC system.
The unconstrained predictions from Section~3.3 must satisfy a doubly constrained structure: on the demand side, each origin $i$ has a finite candidate pool $E_i$ (Section~3.3), ensuring that $\sum_j a_{ij} \leq E_i$; on the supply side, each ESC school $j$ has a finite number of contracted slots $K_j$, ensuring that $\sum_i a_{ij} \leq K_j$.
Because multiple ESC schools draw from overlapping origin catchment areas, accepting students along one path reduces availability for others---a coupled depletion structure that cannot be resolved by independent per-school calculations.
We formalize this as a stochastic allocation process. For a given scenario $s$ and random seed $r$, let $\pi^{(r)}$ be a uniformly random permutation of all origin-destination pairs in the candidate beneficiary pool $\mathcal{P}$.
We process pairs sequentially in the order $\pi^{(r)}$, maintaining residual capacities $e_i$ (initialized to $E_i$) and $k_j$ (initialized to $K_j$).
For each pair $(i,j)$, the accepted flow is:
\begin{equation}\label{eq:acceptance}
  a_{ij}^{(s,r)} = \min\!\big(e_i,\; k_j,\; \hat{y}_{ij}^{(s)}\big)
\end{equation}
after which both residual capacities are updated: $e_i \leftarrow e_i - a_{ij}^{(s,r)}$ and $k_j
\leftarrow k_j - a_{ij}^{(s,r)}$.
The iteration terminates early when either total remaining candidates or total remaining slots reaches zero.
The random permutation $\pi^{(r)}$ determines which paths are prioritized when capacity is scarce; by averaging over $R = 100$ independent seeds ($r = 0, \ldots, 99$), we obtain Monte Carlo means and 95\% confidence intervals that capture the allocation uncertainty due to processing order.
The total simulated enrollment at ESC school $j$ under scenario $s$ and seed $r$ is $Y_j^{(s,r)} = \sum_i a_{ij}^{(s,r)}$.

\paragraph{Decongestion Attribution.}

Not all ESC enrollment contributes to public school decongestion---only students diverted from origins that also feed congested public schools.
Let $\mathcal{G}$ denote the set of congested public junior high schools and let $\mathcal{O}^* = \{i : \exists\, g \in \mathcal{G} \text{ s.t.\ } F_{ig} > 0\}$ be the set of \emph{congested-feeding origins}, i.e., elementary schools whose graduates are observed to flow to at least one school in $\mathcal{G}$.
For each ESC school $j$, we define the congested fraction:
\begin{equation}\label{eq:congested-frac}
  \phi_j = \frac{\sum_{i \in \mathcal{O}^*} \hat{y}_{ij}}{\sum_{i} \hat{y}_{ij}}
\end{equation}
representing the share of $j$'s predicted demand originating from congested-feeding schools, computed from unconstrained model predictions as a pre-simulation approximation.
We then define two complementary decongestion measures for each ESC school $j$.
The \emph{total predicted decongestion role} captures the full congestion-relevant simulated flow:
\begin{equation}\label{eq:total-decongestion}
  D_j^{\mathrm{total}(s,r)} = Y_j^{(s,r)} \cdot \phi_j
\end{equation}
The \emph{marginal decongestion potential} isolates untapped capacity beyond current observed enrollment
$b_j = \sum_i F_{ij}$:
\begin{equation}\label{eq:marginal-decongestion}
  D_j^{\mathrm{marg}(s,r)} = \max\!\big(0,\; Y_j^{(s,r)} - b_j\big) \cdot \phi_j
\end{equation}
The total measure quantifies the ESC system's structural role in absorbing students who would otherwise attend congested public schools; the marginal measure isolates the additional relief achievable if all contracted slots were fully utilized.

\paragraph{Existing and Hypothetical Path Decomposition.}

The augmented candidate beneficiary pool (Section~3.3) includes both historically observed pathways and counterfactual pairings identified through a lexicographical search of the school network.
We classify each pair $(i,j) \in \mathcal{P}$ as \emph{existing} if at least one beneficiary was observed on that route ($F_{ij} > 0$), and \emph{hypothetical} otherwise.
Let $\mathcal{E} \subset \mathcal{P}$ denote existing paths and $\mathcal{H} = \mathcal{P} \setminus
\mathcal{E}$ the hypothetical paths.
We decompose each school's simulated enrollment accordingly:
\begin{equation}\label{eq:decomposition}
  Y_j^{(s,r)} = \underbrace{\textstyle\sum_{i:(i,j) \in \mathcal{E}} a_{ij}^{(s,r)}}_{\text{existing}}
\;+\; \underbrace{\textstyle\sum_{i:(i,j) \in \mathcal{H}} a_{ij}^{(s,r)}}_{\text{hypothetical}}
\end{equation}
This decomposition carries through to both decongestion measures and distinguishes gains achievable by scaling up current enrollment patterns from those requiring the activation of new routes not yet observed in practice.

\paragraph{Counterfactual Scenarios.}

We evaluate five counterfactual scenarios corresponding to net cost reductions of \{1K, 5K, 10K, 15K, 20K\} Philippine pesos per student, which modify the cost term $c_{ij}$ in the gravity model (Section~3.2) and thereby adjust the predicted flows $\hat{y}_{ij}^{(s)}$.
The slot capacity $K_j$ and candidate pools $E_i$ are held constant across scenarios, isolating the demand-side effect of subsidy changes under fixed supply constraints.
This design allows us to disentangle whether the binding constraint on ESC-mediated decongestion is the price of access or the physical supply of school places.

\section{Results}

\subsection{Student Flow Modeling Results}

Table \ref{tab:nbr_results} presents the estimation result of the gravity model using negative binomial regression.
The model employs a log-log specification for distance and cost, allowing coefficients to be interpreted as elasticities.

\begin{table}[ht]
    \caption{Negative Binomial Regression Results}
    \label{tab:nbr_results}
    \centering
    \setlength{\tabcolsep}{1mm}
    \begin{tabular}{@{}lcccc@{}}
        \toprule
        \textbf{Variable} & \textbf{Coef.} & \textbf{Std. Error} & \textbf{$z$-value} & \textbf{$p$-value} \\
        \midrule
        Intercept                           & 3.3944  & 0.130 &  26.121 & 0.000 \\
        $\ln(\text{distance})$              & -0.4509 & 0.010 & -46.488 & 0.000 \\
        $\ln(\text{net cost})$              & -0.1004 & 0.013 & -7.582  & 0.000 \\
        ESC School Rating                   & -0.0204 & 0.008 & -2.503  & 0.012 \\
        $\ln(\text{origin LGU income})$     & -0.0207 & 0.004 & -5.053  & 0.000 \\
        $\ln(\text{dest. LGU income})$      & -0.0489 & 0.004 & -11.406 & 0.000 \\
        \midrule
        \multicolumn{5}{l}{\textit{Origin Region FE (Reference: NCR)}} \\
        \quad Region III                    & -0.0205 & 0.026 & -0.781  & 0.435 \\
        \quad Region IV-A                   & -0.0500 & 0.021 & -2.438  & 0.015 \\
        \midrule
        \multicolumn{5}{l}{\textit{Destination Region FE (Reference: NCR)}} \\
        \quad Region III                    & -0.0237 & 0.026 & -0.911  & 0.362 \\
        \quad Region IV-A                   &  0.0178 & 0.019 &  0.921  & 0.357 \\
        \midrule
        Dispersion ($\alpha$)               &  0.3925 & 0.011 &  35.737 & 0.000 \\
        \midrule
        \multicolumn{5}{l}{\textit{Observations: 29,224}} \\
        \multicolumn{5}{l}{\textit{Pseudo $R^2$ (McFadden): 0.08625}} \\
        \multicolumn{5}{l}{\textit{Log-Likelihood: -53,592}} \\
        \bottomrule
    \end{tabular}
    \smallskip
    \begin{minipage}{\linewidth}
        \small
        \textit{Note:} Standard errors are clustered at the origin school level. Distance and cost are log-transformed, representing elasticities. All income variables are LGU-level revenues.
    \end{minipage}
\end{table}

\textbf{Distance Elasticity.}
The distance decay effect is highly significant.
This implies that a 10\% increase in road distance between origin and destination schools results in approximately a 4.5\% decrease in expected student flows.
Equivalently, doubling the distance reduces expected enrollment by approximately 27\% ($= 1 - 2^{-0.45}$).
Distance has the strongest effect in the model, indicating that geographic accessibility is the primary barrier to private school enrollment among subsidy-eligible families.

\textbf{Cost Elasticity.}
Net enrollment cost ($\text{tuition} - \text{subsidy}$) also exhibits a statistically significant negative effect.
This indicates that a 10\% reduction in out-of-pocket tuition cost increases expected student flows by approximately 1.1\%.
Doubling the net cost would decrease enrollment by roughly 7.4\% ($= 0.5^{-0.11} - 1$).
Because the magnitude of this cost elasticity is notably smaller than the distance elasticity (by nearly a factor of 4), it can be said that subsidies alone have limited effectiveness in driving enrollment shifts if students lack access to nearby private schools.
This suggests that geographic expansion of ESC-participating schools may be more impactful than purely increasing subsidy amounts for geographically constrained families.

\textbf{Destination School Quality.}
The ESC rating coefficient indicates that higher-rated schools receive fewer ESC beneficiaries.
A post-hoc analysis revealed that higher-rated schools charge premium tuition that exceeds ESC subsidy levels, making them financially inaccessible despite higher ratings.
Alternatively, top-rated schools may also be capacity-constrained and selective, limiting ESC beneficiary intake.
A third possibility is that the ESC rating, which is based on DepEd compliance and administrative criteria, may not align with quality dimensions families prioritize, such as proximity, transportation availability, or peer composition.
The small magnitude ($\beta = -0.02$) suggests this effect, while statistically significant, has limited practical importance compared to distance and cost factors.

\textbf{Origin LGU Income Effects.}
The scaled origin income coefficient is negative and highly significant ($\beta = -0.0524$, $p < 0.001$), indicating that students from higher-income municipalities are less likely to utilize the subsidy program.
This could indicate that the ESC program is successfully reaching its intended beneficiaries, the students from lower-income communities where public school quality gaps are most pronounced.
The magnitude of this income effect is comparable to the cost elasticity, suggesting that origin municipality wealth plays a substantive role in enrollment decisions.

\textbf{Destination LGU Income Effects.}
The negative coefficient for destination LGU income suggests that students avoid schools in the wealthiest municipalities.
This implies that hidden costs, such as daily meals, school supplies, and other miscellaneous costs act as a barrier even when tuition is subsidized.
Students from modest backgrounds may avoid schools in wealthy areas because of the financial burden of higher cost-of-living standards.

\textbf{Regional Fixed Effects.}
The regional fixed effects reveal substantial spatial heterogeneity in subsidy
participation patterns.
On the demand side, students from Region III and Region IV-A exhibit lower participation per origin-destination pair compared to NCR students, controlling for distance, cost, and income.
However, in absolute terms, Region IV-A generates the largest share of ESC beneficiaries (44.5\% of total flows), reflecting its larger population base.

On the supply side, destination schools in Region III and Region IV-A enroll more ESC students per school than NCR schools, all else equal.
This pattern manifests in aggregate flows, where NCR experiences a net outflow of students while Region IV-A  receives a net inflow.
This demand-supply mismatch suggests that NCR private schools face binding capacity constraints or strategically limit ESC participation despite higher per-capita demand from NCR families, forcing students to seek ESC slots in neighboring regions at additional cost.

\textbf{Overdispersion}
The estimated dispersion parameter ($\alpha = 0.3948$, $p < 0.001$) confirms substantial overdispersion in the data, with variance significantly exceeding the mean.
This validates the use of negative binomial regression over standard Poisson, ensuring appropriate handling of flow heterogeneity.

\subsection{Policy Simulation Results}

Drawing on the empirical finding that geographic proximity serves as a significantly stronger driver of school choice than tuition cost, we evaluate the system-wide impact of these behavioral elasticities by integrating our stochastic gravity model predictions into a constrained spatial allocation framework, and measure the resulting decongestion under five subsidy scenarios.

Table~\ref{tab:simulation_results} reports the Monte Carlo simulation results across all scenarios, with
observed ESC beneficiary enrollment as a baseline.
Two comparisons are presented: the gap between observed and simulated enrollment (reflecting untapped slot
capacity), and the change across subsidy scenarios (reflecting the demand-side effect of cost
reductions).

\begin{table}[ht]
  \caption{Constrained Simulation Results Across Subsidy Scenarios}
  \label{tab:simulation_results}
  \centering
  \setlength{\tabcolsep}{1.5mm}
  \begin{tabular}{@{}lrrrr@{}}
      \toprule
      \textbf{Scenario} & \textbf{Net Cost} & \textbf{Predicted} & \textbf{$\Delta$ from} &
\textbf{$\Delta$ from} \\
       & \textbf{Reduction} & \textbf{Flow $\bar{Y}^{(s)}$} & \textbf{Observed} & \textbf{$-$1K} \\
      \midrule
      Observed ($b_j$) & --- & 74,232 & --- & --- \\
      \midrule
      $-$1K   & \textpeso 1,000  & 99,992  & $+$34.7\% & ---       \\
      $-$5K   & \textpeso 5,000  & 100,442 & $+$35.3\% & $+$0.4\%  \\
      $-$10K  & \textpeso 10,000 & 100,944 & $+$36.0\% & $+$1.0\%  \\
      $-$15K  & \textpeso 15,000 & 101,428 & $+$36.6\% & $+$1.4\%  \\
      $-$20K  & \textpeso 20,000 & 101,818 & $+$37.2\% & $+$1.8\%  \\
      \bottomrule
  \end{tabular}
  \smallskip
  \begin{minipage}{\linewidth}
      \small
      \textit{Note:} Predicted flow is the Monte Carlo mean over $R = 100$ iterations. System-level
standard deviation is $\sim$11 students ($<$0.01\% of mean), indicating stable convergence. Observed
baseline is total ESC beneficiaries from student flow records. All scenarios use the same slot capacity
$K_j$ and candidate pool $E_i$.
  \end{minipage}
\end{table}

\textbf{The slot ceiling ddominates price sensitivity.}

The gap between observed enrollment and simulated enrollment is substantial.
Under even the most conservative scenario ($-$1K), the model predicts approximately 26,000 additional students would fill ESC slots if all contracted capacity were utilized, a 34.7\% increase over current enrollment.
Yet varying the subsidy from a 1,000-peso to a 20,000-peso net cost reduction increases predicted flow by only 1,826 students, or 1.8\%.
This near-flat response is a direct consequence of the structural imbalance in the system: 562,958 candidates compete for 106,654 contracted slots, a demand-to-supply ratio of 5.3$\times$.
At this level of oversubscription, reducing price reshuffles which students fill slots but barely changes how many slots are filled.

\textbf{Decongestion is geographically concentrated.}
We report the decongestion decomposition for the $-$1K scenario, which is representative of all scenarios given the minimal cross-scenario variation.
Of the 99,992 predicted enrollees, 95,695 (95.7\%) flow through congested-feeding origins---elementary schools whose graduates also attend congested public junior high schools.
The marginal decongestion potential, $\sum_j D_j^{\mathrm{marg}}$, totals 24,820 students, of which 98.1\% originates from congested-feeding schools.
In other words, virtually all of the ESC system's untapped absorption capacity is geographically positioned to relieve the most overcrowded public schools.
Among the 1,373 ESC schools in the system, 1,057 exhibit positive marginal capacity (predicted enrollment exceeding current enrollment), while 284 show negative marginals---of which 185 are structurally over-enrolled (current enrollment already exceeds contracted slots) and 99 reflect model under-prediction.

\textbf{Over Half of Untapped Potential Lies in Unobserved Pathways.}
The existing-hypothetical decomposition (Equation~\ref{eq:decomposition}) reveals that 53.9\% of marginal decongestion flows through hypothetical paths---origin-destination pairs where no beneficiary is currently observed.
The remaining 46.1\% flows through existing paths that the model predicts could absorb additional students beyond current levels.
This split suggests that realizing the full decongestion potential of the ESC system requires not only expanding enrollment along established routes but also activating new pathways between congested-feeding origins and nearby ESC schools that students have not yet considered.

\textbf{Decongestion is meaningful but bounded.}
To contextualize the magnitude of the ESC system's decongestion role, we compare it against the scale of public school congestion.
A total of 416,458 students currently flow to the 1,254 congested public junior high schools, representing 23.3\% of all Grade~7 transitions.
The model's total predicted congestion-relevant flow of 95,695 represents a substantial structural counterweight to this pressure.
However, the marginal component---the 24,820 additional students absorbable through full slot utilization---accounts for approximately 6\% of the total congested flow.
Full utilization of existing ESC capacity would meaningfully reduce pressure on the most crowded public schools but would not, on its own, resolve the congestion problem.

These findings indicate that the primary policy lever for ESC-mediated decongestion is not the subsidy amount but the supply of school places.
Expanding the number of contracted slots, or reallocating existing slots from surplus to deficit areas, would directly raise the slot ceiling that currently binds the system.
However, these projections assume that all contracted capacity is operationally accessible, an assumption we revisit in the next section.

\section{Limitations and Ethical Considerations}

Our behavioral estimation assumes that historical enrollment patterns serve as a reliable proxy for future preferences. This assumption does not account for temporal shifts in school reputation or sudden socioeconomic shocks that may occur between academic years, which could significantly alter the predictive accuracy of the model. Furthermore, while our lexicographical redistribution algorithm mirrors the hierarchy of choice observed in our negative binomial results—prioritizing geographic proximity and net tuition cost—real-world household decision-making often involves qualitative factors not captured in administrative datasets. These factors include religious affiliation, specialized curricula, or family legacy, all of which may override the economic and spatial friction modeled in our gravity framework.

A critical operational limitation of the current redistribution algorithm is the assumption that private institutions can instantaneously absorb students up to their authorized capacity. In practice, schools may face internal bottlenecks, such as teacher shortages or classroom maintenance, that temporarily reduce their functional capacity below the officially recorded limit. Consequently, our computational framework should be treated as a decision support tool for human policymakers rather than an automated assignment system. We emphasize that these simulations must be integrated with rigorous auditing and verification of school facilities to ensure the integrity of simulated outcomes, allowing for manual overrides based on ground-level intelligence.

Looking forward, a future direction for this research involves the dynamic calibration of subsidy slots. While this study treats the ESC slot ceiling as a fixed institutional constraint, future iterations could transition toward a responsive system that adjusts functional capacity based on real-time operational readiness and facility audit data. Additionally, the gravity model’s predictive engine could be used to recommend the optimal geographic redistribution of authorized slots, ensuring that private school capacity is strategically expanded or reallocated to high-congestion corridors where the behavioral pull is strongest.

This study was conducted under a formal institutional partnership with the Government Assistance and Subsidies Service Office of the Department of Education. To ensure strict compliance with the Philippine Data Privacy Act of 2012, all sensitive administrative records were de-identified and aggregated into origin-destination pairs prior to analysis. No personally identifiable information (PII) was accessed or stored during the modeling process, ensuring that the pursuit of system-wide visibility remains consistent with our mandate to protect student privacy.

\section{Conclusion}

This work establishes a computational policy framework for addressing school congestion in the Philippine basic education system through data-driven student flow modeling and constrained spatial allocation.
By harmonizing fragmented administrative records across multiple government agencies with open-source geospatial data, we constructed the first comprehensive representation of nationwide student mobility patterns, enabling systematic evaluation of subsidy interventions that was previously operationally infeasible.

Three empirical patterns fundamentally reshape how the Department of Education should approach decongestion policy.
First, behavioral elasticities reveal that geographic accessibility constrains choice more severely than tuition barriers.
Second, supply constraints bind the system more tightly than demand-side incentives.
Contracted slot capacity limits absorption to 106,654 students despite 562,958 eligible candidates, and increasing subsidy amounts from ₱1,000 to ₱20,000 yields marginal enrollment gains of only 1.8\%, while full slot utilization could absorb 34.7\% more students.
Third, untapped decongestion potential concentrates in unactivated pathways: 53.9\% of marginal capacity flows through origin-destination pairs where no beneficiary currently exists, suggesting that strategic outreach and slot reallocation could yield substantial relief without additional infrastructure expenditure.

These findings impose architectural constraints on policy design.
Subsidy programs cannot overcome spatial mismatches between congested public schools and available private capacity.
That 95.7\% of the predicted enrollees flow through congested-feeding origins, yet the marginal impact is only 6\%, confirms that demand-side interventions alone cannot resolve systemic overcrowding.
Supply-side expansion through targeted slot increases in high-congestion areas represents the most effective intervention.

\section{GenAI Disclosure}
The authors used large language models to assist with grammar checking in the manuscript and code debugging.
All AI-generated content was reviewed, verified, and revised by the authors.
The authors take full responsibility for the accuracy and integrity of the research.

\bibliographystyle{ACM-Reference-Format}
\bibliography{references}

\appendix

\section{Model Comparison with OLS and Poisson}
\label{app:model_comparison}

To further justify the selection of the Negative Binomial Regression framework, we compared its performance against two traditional benchmarks: an Ordinary Least Squares (OLS) model on log-transformed flows and a standard Poisson regression.

\begin{table}[h]
    \caption{Comparison of Count-Based Models}
    \label{tab:model_comparison}
    \centering
    \begin{tabular}{lcc}
    \toprule
    \textbf{Metric} & \textbf{Poisson Regression} & \textbf{Negative Binomial} \\
    \midrule
    Log-Likelihood & $-64{,}568.73$ & $\mathbf{-53{,}653.45}$ \\
    AIC            & $129{,}155.47$ & $\mathbf{107{,}326.91}$ \\
    Dispersion ($\alpha$) & Fixed at 0 & $0.3948^{***}$ \\
    \bottomrule
    \end{tabular}
    \smallskip

    \begin{minipage}{\linewidth}
    \small
    \textit{Note:} $^{***}$ indicates significance at the $p < 0.001$ level. The
    significantly higher log-likelihood and lower AIC for the NBR model indicate
    superior fit for the overdispersed flow data.
    \end{minipage}
\end{table}

\subsection{Jensen's Inequality}

Traditionally, gravity models were estimated using OLS:

\begin{equation}
    \ln(F_{i,j}) = \beta_0 + \beta_1 \ln(d_{i,j}) + \dots + \epsilon_{i,j}
\end{equation}

However, as established by \cite{silva:2006}, this approach is structurally flawed for spatial interaction data.
First, the implications of Jensen's inequality $E(\ln y) \neq \ln E(y)$ mean than OLS estimates are biased and inconsistent in the presence of heteroskedasticity, which is present in our data.
Second, log-linearization requires the exclusion or ad hoc adjustment of zero-flow observations, which truncates the dataset and further biases elasticity estimates.
Finally, OLS is a continuous estimator that fails to account for the discrete nature of student transitions.
By treating student flows as non-negative integers rather than continuous values, the NBR framework more accurately reflects the underlying data-generating process of individual enrollment decisions.
It is possible that the OLS model will yield a lower AIC (in fact, it does), but this is a mathematical artifact of the logarithmic transformation that reduces the scale of the dependent variable and its associated residuals.
Consequenely, the OLS likelihood is not directly comparable to count-based models (Poisson and NBR), and we reject it on theoretical grounds in favor of a framework that preserves the original scale of student flows.

\subsection{Overdispersion in Poisson Models}

While the Poisson model uses a log-link function to avoid the bias of OLS, it assumes equidispersion, where $Var(y) = E(y)$.
In the context of the Philippine basic education system, student transitions exhibit a power-law distribution where certain schools attract massive inflows while others only attract a few.
This leads to overdispersion, where the variance significantly exceeds the mean. As shown in Table~\ref{tab:model_comparison}, moving from a Poisson to an NBR specification results in a massive improvement in log-likelihood (from -64,568 to -53,653).
Furthermore, the estimated dispersion parameter ($\alpha = 0.3948$) is highly significant ($p < 0.001$), formally rejecting the Poisson model and confirming the NBR as the only robust choice for this dataset.

\section{Model Specification}
\label{app:model_specification}

To determine the optimal configuration for the predictive framework, we conducted a hierarchical stepwise evaluation.
This process involves starting with a parsimonious gravity model and incrementally adding school-quality attributes, regional fixed effects, and socioeconomic controls.
We evaluate each specification using three criteria:
\begin{enumerate}
    \item Information Criteria (AIC/BIC): Lower values indicate a better balance between model fit and complexity.
    \item Likelihood Ratio Test (LRT):  A $\chi^2$ test to determine if the improvement in Log-Likelihood from adding new variables is statistically significant.
    \item Dispersion Parameter ($\alpha$): Monitoring the overdispersion to ensure the negative binomial specification remains appropriate as controls are added.
\end{enumerate}

\begin{table}[h]
    \caption{Hierarchical Model Comparison and Specification Tests}
    \label{tab:hierarchy}
    \centering
    \resizebox{\columnwidth}{!}{
    \begin{tabular}{lcccccc}
    \toprule
    \textbf{Model} & \textbf{Log-Lik.} & \textbf{AIC} & \textbf{BIC} & \textbf{LRT Stat} & \textbf{p-value} & \textbf{$\alpha$} \\
    \midrule
    0. Baseline (Dist. + Cost) & $-53{,}838.5$ & $107{,}684.9$ & $107{,}718.0$ & --- & --- & 0.4022 \\
    1. + School Rating & $-53{,}837.2$ & $107{,}684.4$ & $107{,}725.8$ & 2.49 & 0.115 & 0.4021 \\
    2. + Origin Region & $-53{,}800.1$ & $107{,}614.3$ & $107{,}672.3$ & 74.13 & $<$ 0.001 & 0.4014 \\
    3. + Dest. Region & $-53{,}790.0$ & $107{,}598.0$ & $107{,}672.6$ & 20.28 & $<$ 0.001 & 0.4007 \\
    4. + Origin Income Only & $-53{,}653.5$ & $107{,}326.9$ & $107{,}409.7$ & 273.11 & $<$ 0.001 & 0.3948 \\
    5. + Dest. Income Only & $-53{,}653.5$ & $107{,}326.9$ & $107{,}409.7$ & 0.00 & 1.000 & 0.3948 \\
    6. + Dual Income & $-53{,}592.4$ & $107{,}206.8$ & $107{,}297.9$ & 122.08 & $<$ 0.001 & 0.3925 \\
    \bottomrule
    \end{tabular}
    }

    \smallskip
    \begin{minipage}{\linewidth}
    \footnotesize
    \textit{Note:} LRT statistics compare each model to the specification in the
    immediately preceding row. Model 6 is the final specification used for policy
    simulations.
    \end{minipage}
\end{table}

The results are presented in Table~\ref{tab:hierarchy}.
Our selection of Model 6 as the final specification is informed by the following observations:

\begin{itemize}
    \item The inclusion of regional fixed effects (Models 2 and 3) significantly improved model fit, confirming that student flows are largely shaped by local school networks and regional boundaries.
    While the school rating (Model 1) was statistically significant on its own, it was retained to control for perceived school quality.
    \item Adding the income of the local government unit (LGU) of the origin school provided the largest jump in log-likelihood, while adding that of the destination school yielded no statistical gain.
    \item The simultaneous inclusion of the LGU income of both origin and destination resulted in a significant improvement.
    This means that while origin income influences a student's starting capacity, the negative coefficient for destination income suggests that students often avoid schools that are economically out of reach, even when tuition is subsidized.
\end{itemize}

\section{Model Validation and Robustness}
\label{app:model_validation}

To evaluate the reliability of our expected flow predictions, we performed a non-parametric cluster bootstrap ($n=1,000$) on the evaluation metrics.
By resampling at the origin-school level, we account for the correlated error structures inherent in our administrative data.
The narrow confidence intervals for our primary friction variables confirm that our policy simulations are grounded in consistent student behaviors, rather than being sensitive to localized data fluctuations.

\begin{table}[h]
    \centering
    \begin{tabular}{@{}lcc@{}}
    \toprule
    \textbf{Metric} & \textbf{Bootstrap Median} & \textbf{95\% CI} \\
    \midrule
    MAE & 1.5980 & [1.5602, 1.6359] \\
    RMSE & 3.8782 & [3.6566, 4.0998] \\
    \bottomrule
    \end{tabular}
\end{table}

We also assess the stability of our flow determinants by re-estimating the chosen model (Model 6 in~\ref{app:model_specification}) across 1,000 bootstrap iterations.
This ensures that our behavioral elasticities—specifically distance and net cost—are representative of the entire school system and not driven by influential outliers.

\begin{table}[h]
    \centering
    \small
    \begin{tabular}{@{}lccc@{}}
    \toprule
    \textbf{Variable} & \textbf{Lower 2.5\%} & \textbf{Median} & \textbf{Upper 97.5\%} \\
    \midrule
    Intercept & 3.1406 & 3.3946 & 3.6407 \\
    $\ln(\text{distance})$ & $-0.4693$ & $-0.4506$ & $-0.4292$ \\
    $\ln(\text{net cost})$ & $-0.1260$ & $-0.1007$ & $-0.0762$ \\
    ESC School Rating & $-0.0372$ & $-0.0207$ & $-0.0050$ \\
    $\ln(\text{origin LGU income})$ & $-0.0283$ & $-0.0205$ & $-0.0128$ \\
    $\ln(\text{dest.\ LGU income})$ & $-0.0575$ & $-0.0488$ & $-0.0403$ \\
    \addlinespace[2mm]
    \multicolumn{4}{l}{\textit{Regional Effects (Ref: NCR)}} \\
    \quad Origin: Region III & $-0.0743$ & $-0.0208$ & $0.0291$ \\
    \quad Origin: Region IV-A & $-0.0897$ & $-0.0504$ & $-0.0099$ \\
    \quad Dest: Region III & $-0.0724$ & $-0.0221$ & $0.0252$ \\
    \quad Dest: Region IV-A & $-0.0176$ & $0.0178$ & $0.0573$ \\
    \bottomrule
    \end{tabular}
\end{table}

The cluster bootstrap results confirm the structural integrity of the model, with key determinants of student flow exhibiting high stability across iterations. The 95\% confidence intervals for spatial and economic friction are notably narrow and do not cross zero, providing robust empirical evidence for a non-compensatory hierarchy of priorities where geographic proximity serves as the primary filter.
Furthermore, the confidence interval for destination LGU income remains entirely negative, statistically confirming the existence of hidden cost barriers, such as elevated prices for meals and supplies, that deter subsidy-eligible students from wealthier districts.
Finally, while students from Region IV-A show a significantly lower propensity to participate, the destination fixed effects for both Region III and Region IV-A cross zero, suggesting that the attractiveness of a destination is effectively captured by institutional quality and economic variables rather than a pure regional preference.

\end{document}